\def\year{2020}\relax
\documentclass[letterpaper]{article} 
\usepackage{aaai20}  
\usepackage{times}  
\usepackage{helvet} 
\usepackage{courier}  
\usepackage[hyphens]{url}  
\usepackage{graphicx} 
\usepackage{algorithmic}
\usepackage{algorithm}
\usepackage{amsmath,amssymb}
\usepackage{multirow}
\graphicspath{{figures/}}
\DeclareGraphicsExtensions{.pdf,.png,.jpg,.jpeg}

\title{Byzantine-Robust Federated Learning through Adaptive Model Averaging}

 \author{Luis Mu\~{n}oz-Gonz\'{a}lez, Kenneth T. Co, Emil C. Lupu \\
 Department of Computing, Imperial College London \\
 180 Queen's Gate, SW7 2AZ, London, UK \\
 \{l.munoz, k.co, e.c.lupu\}@imperial.ac.uk 
 }

\begin{document}

\maketitle

\begin{abstract}
Federated learning enables training collaborative machine learning models at scale with many participants whilst preserving the privacy of their datasets. Standard federated learning techniques are vulnerable to Byzantine failures, biased local datasets, and poisoning attacks. In this paper we introduce \emph{Adaptive Federated Averaging}, a novel algorithm for robust federated learning that is designed to detect failures, attacks, and bad updates provided by participants in a collaborative model. We propose a Hidden Markov Model to model and learn the quality of model updates provided by each participant during training. In contrast to existing robust federated learning schemes, we propose a robust aggregation rule that detects and discards bad or malicious local model updates at each training iteration. This includes a mechanism that blocks unwanted participants, which also increases the computational and communication efficiency. Our experimental evaluation on 4 real datasets show that our algorithm is significantly more robust to faulty, noisy and malicious participants, whilst being computationally more efficient than other state-of-the-art robust federated learning methods such as \emph{Multi-KRUM} and \emph{coordinate-wise median}.
\end{abstract}

\section{Introduction}
Distributed learning has emerged as a mechanism to cope with the increasing complexity of machine learning models and the amount of data available to train them. Among existing approaches, federated machine learning has become a popular implementation to build shared machine learning models from a federation of participants \cite{mcmahan2017,konevcny2016,konevcny2016b,shokri2015}. In this approach, there is a central node (server) that controls the learning process and aggregates the information from the participants, commonly referred to as \emph{clients}, which train the model locally using their own local datasets. The clients then send the model updates to the server that aggregates all the information to update the shared model. Federated learning enables building shared machine learning models while keeping the clients' data private. Federated learning can operate in different settings, including cases where the number of clients is large, synchronous and asynchronous operations, or scenarios where the communications between the clients and the server are slow or unstable \cite{konevcny2016b}. 

\cite{blanchard2017} showed that standard federated learning strategies fail in the presence of faulty and malicious clients. Thus, just one bad client can compromise the entire performance and the convergence of the shared model. To mitigate this limitation, different robust federated learning strategies have already been proposed in the literature \cite{blanchard2017,mhamdi2018hidden,yin2018,damaskinos2019,chen2017,chen2018,xie2019}. These techniques have been proven to be robust against Byzantine adversaries, including some poisoning attacks \cite{huang,nelsonSpam}. Some of these techniques rely on robust statistics (e.g. median estimators) for updating the aggregated model \cite{blanchard2017,mhamdi2018hidden,yin2018,chen2017}, which can be computational expensive for large models and number of clients compared to standard aggregation rules such as Federated Averaging \cite{mcmahan2017}. These techniques also disregard the fraction of training data points provided by each client, which can be limiting in cases where clients provide a significantly different amount of data points to train the model. Other approaches such as DRACO \cite{chen2018} require to add some redundancy to achieve robustness and the data provided by the clients may require leaving their facilities to build client nodes with partially overlapping datasets. 

In this paper we introduce \emph{Adaptive Federated Averaging} (AFA) a novel approach for Byzantine-robust federated learning. We propose an algorithm that aims to detect and discard bad or malicious client's updates at every iteration by comparing the similarity of these individual updates to the one for the aggregated model. In contrast to other robust federated algorithms, we propose a method to estimate the quality of the client's updates through a Hidden Markov Model. This allows us to define a mechanism to block faulty, noisy or malicious clients that systematically send bad model updates to the server, reducing the communication burden between the clients and the server and the time required by the server to compute the aggregated model. Our experimental evaluation with 4 real datasets show that our model is more robust to different scenarios including faulty, malicious and noisy adversaries, compared to other state-of-the-art methods such as Multi-KRUM (MKRUM) \cite{blanchard2017} and \emph{coordinate-wise median} (COMED) \cite{yin2018}. We also show empirically that our proposed method is computationally much more efficient than MKRUM and COMED, reducing the time required to update the model in the server at each training iteration.

\section{Related Work}
The first federated optimization algorithms based on the aggregation of clients' gradients were proposed in \cite{shokri2015,konevcny2016,konevcny2016b}. A more communication-efficient implementation with \emph{Federated Averaging} (FA) was described in \cite{mcmahan2017}, where model parameters are updated with a weighted average of the model parameters sent by selected clients at each iteration.

As previous techniques are fragile in the presence of faulty or malicious clients, different robust techniques have been proposed in the research literature. \cite{blanchard2017} propose KRUM, a Byazntine-robust aggregation rule based on the similarity of the gradients or updates provided by all the clients at each iteration. They provide a theoretical analysis that guarantees robustness under certain assumptions. As KRUM converges slowly  compared to other aggregation rules, the authors also introduce MKRUM, a variant of the previous algorithm that achieves similar performance at a faster convergence rate. \cite{chen2017,yin2018} introduce robust aggregation rules based on the computation of the median for the model updates. \cite{mhamdi2018hidden} propose \emph{Bulyan} an recursive algorithm that leverages existing Byzantine-robust aggregation rules, e.g. KRUM or COMED \cite{yin2018}, to improve robustness against poisoning attacks. A framework implementing some of these aggregation rules is presented in \cite{damaskinos2019}. These methods focus on stochastic gradient descent approaches for computing the aggregated model (although they can also be applied for FA-based schemes), which can be very inefficient communication-wise \cite{mcmahan2017}. On the other hand, these approaches assume that the clients have the same (or a similar) number of training data points and, thus, the proposed aggregation rules can be inefficient when some of the clients have significantly more data than others. Finally, the computational complexity for computing these aggregation rules can be demanding, especially if the number of clients or the number of parameters is large. In contrast, our proposed robust aggregation rule is based on FA, i.e. it accounts for the number of data points provided by each client and is more communication-efficient. As we show in our experimental evaluation, the computational complexity is also reduced compared to other robust methods such as MKRUM or COMED. 

\cite{xie2019} propose \emph{Zeno}, a robust aggregation mechanism that aims to rank the clients' gradient estimator. Then, only the gradients of the top $k$ highest rated clients is considered for updating the model at each iteration. However, the number of selected clients, $k$, must be specified in advance. Our proposed approach, in contrast, aims to identify and discard bad (or malicious) client's updates at each iteration, and then, uses the information from all the \emph{good} clients to compute the model update. This removes the reliance on a pre-specified parameter $k$, and allows us to utilize information from all clients, especially in cases where there are few or no bad clients, which would otherwise be discarded in the case of \emph{Zeno}.
 
The research literature in adversarial machine learning \cite{huang} has shown that learning algorithms are vulnerable to poisoning attacks \cite{biggioSVM,mei,luisPoisoning}, where attackers can compromise a fraction of the training dataset to manipulate degrade the algorithms' performance. Some of the robust strategies described previously have been proven vulnerable against more subtle attack strategies, including targeted poisoning attacks \cite{bhagoji2019} and \emph{backdoors} \cite{bagdasaryan}. Additionally, more specific attacks targeting KRUM, COMED and \emph{Bulyan} \cite{baruch2019,xie2019b} have shown that it is possible to compromise the performance of these robust techniques.
\section{Adaptive Federated Averaging}
In typical supervised learning tasks (including classification and regression), given a training dataset with $n$ data points, ${\cal D}_{tr} = \{ x_i,y_i \}_{i=1}^n$, the parameters of the machine learning model $w \in \mathcal{R}^d$ are learned by the minimization of a loss function $\ell$ evaluated on ${\cal D}_{tr}$: 
\begin{equation}
\min_{w \in {\mathcal R}^d} \frac{1}{n} \sum_{i=1}^n \ell(x_i,y_i; w)
\end{equation} In a federated machine learning task we assume there are $K$ clients over which the data is partitioned in $K$ datasets (referred to as \emph{shards}), ${\cal D}_{k} = \{ x_i^k,y_i^k \}_{i=1}^{n_k}$.\footnote{The size of the clients datasets is not necessarily equal.} In this case, we can re-write the previous objective as:
\begin{equation}
\min_{w \in {\mathcal R}^d} \frac{1}{n} \sum_{k=1}^K \frac{n_k}{n} {\cal L}_k (w) 
\label{eqFMLobj}
\end{equation} where ${\cal L}_k (w) = \frac{1}{n_k} \sum_{i=1}^{n_k} \ell(x^k_i,y^k_i; w)$ and $n = \sum_{k=1}^K n_k$. In distributed optimization the shards of the clients are assumed to be IID, whereas federated optimization does not necessarily rely on this assumption. Federated learning algorithms can be trained with gradient descent strategies by updating the parameters of the model from the local updates provided by the clients. Thus, at each iteration $t$, a subset of $K_t \subset K$ clients is selected for synchronous aggregation, i.e. the central node responsible for computing the aggregated model sends the latest update of the model to the selected clients. Then, these clients compute local updates of the model using their own shards (e.g. by computing the gradients of the loss function w.r.t. the model parameters) and send the corresponding local updates to the central node, which updates the model according to a defined aggregation rule \cite{konevcny2016}.
Typical implementations use gradient descent and compute the model updates in the central node as $w_{t+1} \leftarrow w_t - \eta \sum_{k \subset K_t} \frac{n_k}{n'} g_k^t$, where $\eta$ is the learning rate and $g_k^t = \nabla_{w_t} {\cal L}_k (w_t)$, and $n' = \sum_{k \subset K_t} n_k$. However, this approach is not efficient in the communications between the central node and the clients. More efficient implementations can be achieved using \emph{Federated Averaging} (FA) \cite{mcmahan2017}, where the aggregation in the central node is computed as $w_{t+1} \leftarrow \sum_{k \subset K_t} \frac{n_k}{n'} w_{t+1}^k$. In this case, the model updates from each client $w_{t+1}^k$ can be computed locally over several iterations of gradient descent, with $w^k \leftarrow w^k - \eta \nabla_{w^k} {\cal L}_k (w^{k})$. If only one gradient update is computed, FA is equivalent to the previous model. FA updates the model by computing a weighted average of the model updates provided by the different clients, taking into account the number of training points provided by each client. 

However, all these techniques are not fault tolerant and can be very fragile to malicious or faulty clients. Thus, it has been shown that, with just one \emph{bad} client, the whole federated optimization algorithm can be compromised leading to bad solutions or convergence problems \cite{mhamdi2018hidden,blanchard2017}. To overcome these limitations, we propose \emph{Adaptive Federated Averaging} (AFA) an algorithm for federated learning robust to faulty, noisy, and malicious clients. In contrast to FA, our algorithm updates the model considering not only the amount of data provided by each client, but also their reliability, by estimating the probability of each client providing \emph{good} (or useful) model updates. Thus, at each iteration, given a set of selected clients $K_t$ that compute their model updates $w_{t+1}^k$, the central node updates the model as:
\begin{equation}
w_{t+1} \leftarrow \sum_{k \subset K_t^g} \frac{p_{k_t} n_k}{N} w_{t+1}^k
\label{eqAFA}
\end{equation} where $p_{k_t}$ is the probability of client $k$ providing good model updates (according to AFA's aggregation rule) at iteration $t$ and $N = \sum_{k \subset K_t^g} p_{k_t} n_k$. The subset $K_t^g \subset K_t$ contains the clients that provide a good update according to the robust aggregation criterion that we describe next.

\subsection{Model Aggregation}
The algorithm to compute AFA's model aggregation is detailed in Algorithm \ref{alg:Aggregation}. Initially, we compute the model update in (\ref{eqAFA}) with the clients in $G$ (all $K_t$ clients in the first iteration of the algorithm). Then, we compute $s_k$, the similarities of the aggregated model parameters, $w_{t+1}$, with the parameters provided by every client in $G$, $w_{t+1}^k$. In our implementation, we used the \emph{cosine similarity}, which provides a normalized similarity measure in the interval $[-1,1]$ and its computation scales linearly in time with the number of parameters of the machine learning model. Note that other similarity or distance metrics can also be applied (e.g. correlation, euclidean distance).  

We calculate the mean (${\hat \mu}_s$), median (${\bar \mu}_s$) and standard deviation ($\sigma_s$) for the similarities computed for all the good clients in $G$. Assuming that the number of malicious or faulty clients is less than $\lfloor K_t/2 \rfloor$, i.e. more than half of the clients are good, the median ${\bar \mu}_s$ must correspond to the similarity value computed for one of the good clients. Then, by comparing the relative position of ${\bar \mu}_s$ and ${\hat \mu}_s$ we identify the direction to look for potential bad client's updates. Thus, if ${\hat \mu}_s < {\bar \mu}_s$ most of the potential bad clients will have a similarity value $s_k$ lower than ${\bar \mu}_s$. Then, if $s_k < {\bar \mu}_s - \xi \ \sigma_s$ the update from client $k$ is considered as \emph{bad}. The threshold is controlled by a parameter $\xi$. This can be the case of malicious clients that perform attacks aiming to remain stealthy, noisy clients whose shards differ from the data of the majority of the clients and faulty clients that send to the server very low values for $w_{t+1}^k$, compared to the majority of clients. On the other hand, if ${\hat \mu}_s \geq {\bar \mu}_s$, updates are considered as bad if $s_k < {\bar \mu}_s + \xi \ \sigma_s$. This can be the case of strong attacks, such as malicious clients that collude to perform attacks that do not consider strong detectability constraints. Moreover, this scenario also includes faulty clients that provide very large values for $w_{t+1}^k$ compared to the rest, so that $s_k$ will be very close to $1$ (when using the cosine similarity) for the faulty client(s) and smaller for the genuine ones.

\begin{algorithm}
   \caption{AFA robust model aggregation}
   \label{alg:Aggregation}
\begin{algorithmic}
\REQUIRE selected clients $K_t$, $p_{k_t}$, $n_k$, $w_{t+1}^k$, $w_t$, $\xi_0$, $\Delta \xi$, $\eta$
    \STATE $B \leftarrow \{ \emptyset \}$, Bad clients' set 
    \STATE $G \leftarrow \{k : k \in K_t \}$, Good clients' set
    \STATE $R \leftarrow \{1\}$ (Initialize as a non-empty set)
    \STATE $\xi \leftarrow \xi_0$
    \WHILE {$R \neq \emptyset$}
    \STATE $R \leftarrow \emptyset$, $N = \sum_{k \in G} p_{k_t} n_k$
    \STATE $w_{t+1} \leftarrow \sum_{k \subset G} \frac{p_{k_t} n_k}{N} w_{t+1}^k$, with $N = \sum_{k \subset K_t^g} p_{k_t} n_k$
    \FOR {$k \in G$}
    \STATE $s_k = $ similarity($w_{t+1}$,$w_{t+1}^k$)
    \ENDFOR
    \STATE Compute ${\hat \mu}_s$, ${\bar \mu}_s$, $\sigma_s$ for clients in $G$
    \IF {${\hat \mu}_s < {\bar \mu}_s$}
    \FOR {$k \in G$}
    \IF {$s_k < {\bar \mu}_s - \xi \ \sigma_s$}
    \STATE $R \leftarrow R \cup \{k\}$, $G \leftarrow G \backslash \{k\}$
    \ENDIF
    \ENDFOR
    \ELSE
    \FOR {$k \in G$}
    \IF {$s_k > {\bar \mu}_s + \xi \ \sigma_s$}
    \STATE $R \leftarrow R \cup \{k\}$, $G \leftarrow G \backslash \{k\}$
    \ENDIF
    \ENDFOR
    \ENDIF
    \STATE $\xi = \xi + \Delta \xi$, $B \leftarrow B \cup R$
    \ENDWHILE
    \STATE $w_{t+1} \leftarrow \sum_{k \subset G} \frac{p_{k_t} n_k}{N} w_{t+1}^k$, with $N = \sum_{k \subset K_t^g} p_{k_t} n_k$
    \RETURN $w_{t+1}$, $G$, $B$
\end{algorithmic}
\end{algorithm}

As different kind of bad or malicious clients can be present, we repeat the algorithm until no more malicious clients are found (e.g. we can have scenarios with faulty clients and weak attacks). As we iterate, we increase the value of $\xi$ by $\Delta \xi > 0$ to limit the number of false positives, i.e. good clients' updates classified as bad. Finally, the algorithm returns the aggregated model update $w_{t+1}$ computed with the set of good clients in $G$. The list with good and bad clients in $G$ and $B$ is also used to estimate $p_{k_{t+1}}$ with the Bayesian model we explain next.

The complexity to compute the aggregation rule in (\ref{eqAFA}) is ${\cal O}(K_t d)$ in time, where $d$ is the number of parameters of the model. The computation of the similarity of the client's updates w.r.t. the aggregated model also scales in time as ${\cal O}(K_t d)$ in the case of using the cosine similarity or the euclidean distance. This contrasts to KRUM \cite{blanchard2017}, where similarities are computed across all clients, scaling as ${\cal O}(K_t^2 d)$. Finally, computing ${\bar \mu}_s$, the median of the client's similarities is ${\cal O}(K_t)$ on average when using efficient algorithms such as \emph{Quickselect} \cite{quickSelect}. This is much more efficient that other median-based aggregation rules such as COMED \cite{yin2018}, which requires computing the median for all the model parameters scaling in time, on average, as ${\cal O}(K_t d)$.

\begin{figure}
\centering
\includegraphics[width=7cm]{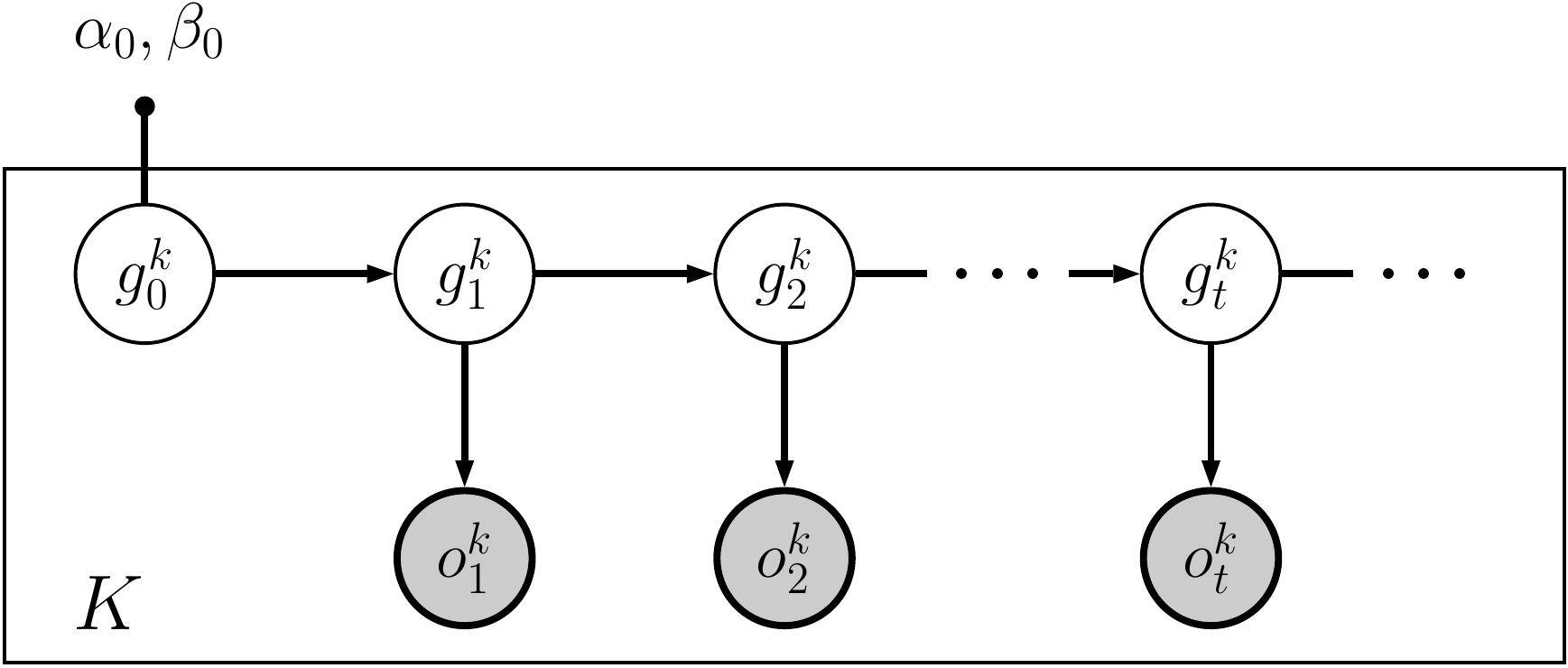}
\caption{AFA Bayesian model to estimate the probability of the clients providing good model updates.}
\label{FigAFAmodel}
\end{figure}

\subsection{Learning the Quality of the Client's Updates}
We propose a Bayesian model to estimate the clients' probability to provide good updates to the model based on the outcome of the aggregation rule in Algorithm \ref{alg:Aggregation} across all iterations. These probabilities are used to compute (\ref{eqAFA}). We model the ability of the clients to provide good model updates as a Hidden Markov Model, as depicted in Fig.~\ref{FigAFAmodel}. Thus, $g_t^k$ is an unobserved random variable that represents the probability of client $k$ providing good updates at iteration $t$. The observed variables $o_t^k$ are binary random variables with the outcome of Algorithm \ref{alg:Aggregation} for client $k$ at iteration $t$, i.e. these variables indicate if client $k$ provided a good update according to the aggregation algorithm. Finally, $g_0^k$ represents the prior belief at $t = 0$. We model this prior as a Beta distribution with parameters $\alpha_0 > 1$, $\beta_0 > 1$. In our experiments we set $\alpha_0 = \beta_0$, so that the expected prior probability for every client is $\mathbb{E} \{ G_0^k \} = \alpha_0 / (\alpha_0 + \beta_0) = 0.5$. The value of $\alpha_0$ and $\beta_0$ also determines the variance of the prior, i.e. larger values of $\alpha_0$ and $\beta_0$ imply a smaller variance. This can be used to control the responsiveness of the model, i.e. the changes in the posterior will be reduced for a smaller variance in the prior. The posterior distribution $p(g_t^k | o_{1:t}^k)$ can be computed through recursive Bayesian estimation, so that:
\begin{equation}
p(g_t^k | o_{1:t}^k) = \frac{p(o_t^k|g_t^k) \ p(g_t^k | o_{1:t-1}^k)}{p(o_t^k | o_{1:t-1}^k)}
\label{eqPosterior}
\end{equation} The likelihood $p(o_t^k|g_t^k)$ can be modelled as a Bernoulli distribution, whereas $p(g_t^k | o_{1:t-1}^k)$ is a Beta distribution with parameters $\alpha_{t-1} = \alpha_0 + n_{g_{t-1}}^k$, $\beta_{t-1} = \beta_0 + n_{b_{t-1}}^k$, where $n_{g_{t-1}}^k$ and $n_{b_{t-1}}^k$ are the number of good and bad updates provided by client $k$ until iteration $t-1$. The posterior also results in a Beta distribution with parameters $\alpha_t^k = \alpha_0 + n_{g_{t}}^k$ and $\beta_t^k = \beta_0 + n_{b_{t}}^k$. 
Then, at each iteration we can compute the probability of client $k$ providing good updates, $p_{k_t}$, used in (\ref{eqAFA}) to update the model at each iteration, as the expectation of the posterior $p(g_t^k | o_{1:t}^k)$: 
\begin{equation}
p_{k_{t+1}} = \mathbb{E} \{ G_t^k | O_{1:t}^k \} = \frac{\alpha_t^k}{\alpha_t^k + \beta_t^k}
\label{eqExpectedProb}
\end{equation}
For simplicity, in the model depicted in Fig.~\ref{FigAFAmodel} we assume that all the clients are used at every iteration. However, the application to the case where only a subset of clients is selected is straightforward. In this case, at every iteration, we just need to update the posterior of the selected clients, whereas the posterior for the non-selected clients remains unchanged. 

\subsection{Blocking Bad Clients}
Robust aggregation methods for federated learning such as \cite{chen2017,blanchard2017,yin2018,mhamdi2018hidden} do not include explicit mechanisms to detect bad or malicious clients. Thus, despite the robustness of these models to Byzantine adversaries, they can be inefficient in terms of computation and communication resources, as they allow bad clients to keep sending useless model updates. In contrast, AFA allows us to detect clients providing bad updates at every iteration (see Algorithm \ref{alg:Aggregation}). The Bayesian model in Fig.~\ref{FigAFAmodel} enables the characterization of clients according to the good and bad updates provided across all iterations. Leveraging this information, we propose here a mechanism to block malicious clients. Thus, at every iteration, we compute the posterior $p(g_t^k | o_{1:t}^k)$ using (\ref{eqPosterior}), which results in a Beta distribution with parameters $\alpha_t^k$ and $\beta_t^k$. Then, at iteration $t$, we block a client $k$ if:
\begin{equation}
\textrm{Pr} ( G_t^k | O_{1:t}^k \leq 0.5 ) = \Phi ( 0.5; \alpha_t^k, \beta_t^k )  > \delta 
\label{eqBlock}
\end{equation} where $\Phi(\cdot)$ is the cumulative distribution function of the Beta posterior distribution in (\ref{eqPosterior}) and $\delta$ is an arbitrary threshold. In our experiments we set $\delta = 0.95$.

Once a client is blocked, the server does not select that client anymore during the training process, increasing the efficiency of the federated learning algorithm. For example, in cases where all clients are selected at each iteration, i.e. $K_t = K$, the blocking mechanism in (\ref{eqBlock}) allows to reduce the communication burden between the server and the client nodes, as no information is exchanged between the server and the blocked clients. The minimum number of iterations required to block bad clients is determined by the parameters of the Beta prior $p(g_0^k)$. Thus, for big values of $\alpha_0$ and $\beta_0$, AFA will require more iterations to block the bad clients. However, although small values can help to block clients earlier, benign clients can also be blocked more easily.

\section{Experiments}
We performed our experimental evaluation on 4 datasets: \emph{MNIST} \cite{mnist}, \emph{Fashion-MNIST} (FMNIST) \cite{fmnist}, \emph{Spambase} \cite{spambase}, and CIFAR-10 \cite{cifar}. The characteristics of these datasets are detailed in the supplement. For MNIST and FMNIST we trained Deep Neural Networks (DNNs) wtih $784 \times 512 \times 256 \times 10$. For Spambase we also trained DNNs with $54 \times 100 \times 50 \times 1$. For CIFAR-10 we used VGG-11 \cite{simonyanVGG}, a well-known Convolutional Neural Network (CNN) architecture. For the optimization in the clients we used gradient descent with a batch size of $200$, training the algorithms for $10$ epochs before sending the updates back to the server. All the details about the architecture and the training settings are also detailed in the supplement. For AFA we set $\alpha_0 = \beta_0 = 3$ for the prior on the client's quality, $\xi = 2$ and $\Delta \xi = 0.5$ for computing the threshold in the aggregation rule, and $\delta = 0.5$ as a threshold for blocking bad clients as in (\ref{eqBlock}). We compared the performance of AFA w.r.t. standard Federated Averaging (FA), Multi-KRUM (MKRUM) \cite{blanchard2017} and COMED \cite{yin2018}. In the experiments we set $K_t = K$ for all the algorithms, i.e. all clients are selected to provide mode updates at each iterations (unless they are blocked in the case of AFA).

For each dataset we considered 4 different scenarios: 1) {\bf Normal operation} (\emph{clean}): we assume that all the clients send genuine updates computed on their corresponding shards. 2) {\bf Byzantine clients} (\emph{byzantine}): similar to \cite{blanchard2017}, bad clients send to the server model updates that significantly differ to the ones from the genuine clients. In our case, the faulty or malicious clients compute the following update at each iteration: $w^k_{t+1} \leftarrow w_t + \Delta w^k_t$, where $\Delta w^k_t$ is a random perturbation drawn from a Gaussian distribution with mean zero and isotropic covariance matrix with standard deviation 20. 3) {\bf Label flipping attack} (\emph{flipping}): we performed a label flipping attack where all the labels of the training points for the malicious clients are set to zero. 4) {\bf Noisy clients} (\emph{noisy}): for MNIST, FMNIST and CIFAR-10 we normalized the inputs to the interval $[-1,1]$. In this scenario, for the selected noisy clients we added uniform noise to all the pixels, so that $x \leftarrow x + \epsilon$, with $\epsilon \sim {\cal U}(-1.4,1.4)$. Then we cropped the resulting values again to the interval $[-1,1]$. For Spambase, where we considered binary features (indicating the presence or absence of a specific keyword), we randomly flipped the value of $30\%$ of the features for all the samples in the datasets of the noisy clients.

\begin{table*}
\begin{scriptsize}
\begin{center}
\begin{tabular}{|l|c|c|c|c||c|c|c|c|} 
    \hline
    \multirow{2}{*}{\bf{~MNIST}} & \multicolumn{4}{c||}{\bf{10 Clients}} & \multicolumn{4}{c|}{\bf{100 Clients}}\\
    \cline{2-5} \cline{6-9}
    & \bf{Clean} & \bf{Byzantine} & \bf{Flipping} & \bf{Noisy} & \bf{Clean} & \bf{Byzantine} & \bf{Flipping} & \bf{Noisy}\\  
    \hline
    AFA & $2.80\pm0.12$ & ${\bf 2.99\pm0.12}$ & ${\bf 2.96\pm0.15}$ & ${\bf 3.04\pm0.14}$ & ${\bf 3.82\pm0.09}$ & ${\bf 4.03\pm0.09}$ & ${\bf 3.98\pm0.08}$ & ${\bf 3.97\pm0.08}$ \\
    FA & ${\bf 2.56\pm0.11}$ & $90.03\pm0.59$ & $8.48\pm1.58$ & $3.79\pm0.17$ & ${\bf 3.86\pm0.14}$ & $89.70\pm0.73$ & $90.20\pm0.00$ & $4.81\pm0.09$ \\
    MKRUM & $3.96\pm0.18$ & $4.01\pm0.13$ & $78.98\pm23.39$ & $3.94\pm0.13$ & $4.68\pm0.11$ & $4.63\pm0.14$ & $4.62\pm0.13$ & $4.61\pm0.13$ \\
    COMED & $2.82\pm0.09$ & $3.17\pm0.16$ & $11.55\pm0.93$ & $3.56\pm0.09$ & $4.08\pm0.11$ & $4.21\pm0.13$ & $7.32\pm0.20$ & $4.51\pm0.10$\\
    \hline
\end{tabular}

\vspace{0.1in}
\begin{tabular}{|l|c|c|c|c||c|c|c|c|} 
    \hline
    \multirow{2}{*}{\bf{~FMNIST}} & \multicolumn{4}{c||}{\bf{10 Clients}} & \multicolumn{4}{c|}{\bf{100 Clients}}\\
    \cline{2-5} \cline{6-9}
    & \bf{Clean} & \bf{Byzantine} & \bf{Flipping} & \bf{Noisy} & \bf{Clean} & \bf{Byzantine} & \bf{Flipping} & \bf{Noisy}\\  
    \hline
    AFA & $14.72\pm1.89$ & $14.11\pm1.16$ & ${\bf 15.45\pm1.88}$ & $15.27\pm1.89$ & $15.72\pm2.11$ & $16.08\pm1.88$ & ${\bf 15.99\pm2.00}$ & $17.48\pm1.32$ \\
    FA & ${\bf 13.60\pm1.62}$ & $89.27\pm0.81$ & $24.52\pm11.23$ & $15.55\pm1.80$ & $16.96\pm1.50$ & $89.55\pm1.18$ & $48.46\pm21.44$ & $18.95\pm0.15$ \\
    MKRUM & $14.23\pm0.21$ & $14.28\pm0.27$ & $34.79\pm3.12$ & $14.21\pm0.24$ & $15.05\pm0.36$ & $15.00\pm0.20$ & ${\bf 15.14\pm0.30}$ & ${\bf 15.25\pm0.52}$ \\
    COMED & ${\bf 12.68\pm0.22}$ & ${\bf 13.33\pm0.23}$ & $24.02\pm0.76$ & $14.43\pm0.32$ & $17.05\pm1.37$ & $15.95\pm1.63$ & $18.11\pm0.35$ & $18.38\pm0.13$\\
    \hline
\end{tabular}

\vspace{0.1in}
\begin{tabular}{|l|c|c|c|c||c|c|c|c|} 
    \hline
    \multirow{2}{*}{\bf{~Spambase}} & \multicolumn{4}{c||}{\bf{10 Clients}} & \multicolumn{4}{c|}{\bf{100 Clients}}\\
    \cline{2-5} \cline{6-9}
    & \bf{Clean} & \bf{Byzantine} & \bf{Flipping} & \bf{Noisy} & \bf{Clean} & \bf{Byzantine} & \bf{Flipping} & \bf{Noisy}\\  
    \hline
    AFA & $6.59\pm0.61$ & ${\bf 7.13\pm0.61}$ & ${\bf 7.09\pm0.51}$ & $7.20\pm0.84$ & ${\bf 6.89\pm0.44}$ & $7.55\pm0.41$ & ${\bf 7.06\pm0.45}$ & ${\bf 7.06\pm0.61}$ \\
    FA & $6.13\pm0.30$ & $47.73\pm4.59$ & $14.10\pm6.87$ & $7.55\pm0.58$ & ${\bf 6.94\pm0.53}$ & $44.64\pm4.91$ & $11.75\pm6.41$ & $7.86\pm0.65$ \\
    MKRUM & $8.22\pm0.52$ & $8.30\pm0.32$ & $8.18\pm0.35$ & $8.19\pm0.34$ & $7.96\pm0.48$ & $7.61\pm0.41$ & $8.03\pm0.38$ & $8.01\pm0.43$ \\
    COMED & $6.49\pm0.30$ & ${\bf 6.96\pm0.88}$ & $8.42\pm0.50$ & $6.96\pm0.65$ & $7.71\pm0.43$ & $7.69\pm0.38$ & $10.56\pm0.62$ & $8.53\pm0.45$\\
    \hline
\end{tabular}

\vspace{0.1in}
\begin{tabular}{|l|c|c|c|c||c|c|c|c|} 
    \hline
    \multirow{2}{*}{\bf{~CIFAR}} & \multicolumn{4}{c||}{\bf{10 Clients}} & \multicolumn{4}{c|}{\bf{100 Clients}}\\
    \cline{2-5} \cline{6-9}
    & \bf{Clean} & \bf{Byzantine} & \bf{Flipping} & \bf{Noisy} & \bf{Clean} & \bf{Byzantine} & \bf{Flipping} & \bf{Noisy}\\  
    \hline
    AFA & $31.94\pm0.88$ & ${\bf 33.15\pm0.69}$ & ${\bf 33.14\pm0.81}$ & $31.61\pm0.64$ & $35.66\pm0.58$ & ${\bf 36.83\pm0.41}$ & ${\bf 37.42\pm0.62}$ & $35.68\pm0.57$ \\
    FA & ${\bf 30.11\pm0.39}$ & $90.00\pm0.00$ & $72.75\pm16.39$ & ${\bf 30.01\pm0.43}$ & $35.84\pm0.54$ & $90.00\pm0.00$ & $80.21\pm14.85$ & $35.93\pm0.62$ \\
    MKRUM & $53.34\pm2.90$ & $53.02\pm2.45$ & $58.33\pm2.94$ & $52.95\pm1.11$ & $77.65\pm10.69$ & $71.15\pm3.41$ & $77.54\pm10.79$ & $72.27\pm2.42$ \\
    COMED & $32.61\pm0.46$ & $35.41\pm0.47$ & $43.99\pm4.54$ & $32.62\pm0.37$ & $35.68\pm0.35$ & $37.22\pm0.35$ & $63.87\pm8.29$ & $35.88\pm0.27$\\
    \hline
\end{tabular}
\end{center}
\end{scriptsize}
\caption{Averaged test error over 10 random training data splits for AFA, FA, MKRUM and COMED in different scenarios: clean datasets for all clients, Byzantine adversaries, label flipping attacks, and noisy clients. For each scenario we considered two settings with 10 and 100 clients with 3 and 30 bad/malicious clients respectively. Best results with statistical significance at the $5\%$ level, according to a Wilcoxon Rank-Sum test, are highlighted in bold.}
\label{tabResults}
\end{table*}

\begin{table*}
\begin{center}
    \begin{tabular}{|l|c|c|c||c|c|c|} 
    \hline
    \multirow{2}{*}{\bf{~Dataset}} & \multicolumn{3}{c||}{\bf{10 Clients}} & \multicolumn{3}{c|}{\bf{100 Clients}}\\
    \cline{2-4} \cline{5-7}
    & \bf{Byzantine} & \bf{Flipping} & \bf{Noisy} & \bf{Byzantine} & \bf{Flipping} & \bf{Noisy}\\ 
    \hline
    {\small MNIST} & 100\% (5.0) & 100\% (13.7) & 100\% (5.0) & 100\% (5.0) & 100\% (5.0) & 100\% (9.2) \\ 
    \hline
    {\small FMNIST} & 100\% (5.0) & 100\% (6.6) & 100\% (5.0) & 100\% (5.0) & 100\% (5.0) & 100\% (7.6) \\
    \hline
    {\small Spambase} & 100\% (5.0) & 100\% (5.1) & 100\% (7.4) & 100\% (5.2) & 100\% (5.8) & 100\% (5.6)\\
    \hline
    {\small CIFAR-10} & 100\% (5.0) & 90\% (8.0) & 20\% (32.7) & 100\% (5.0) & 100\% (13.1) & 2.7\% (84.1) \\
    \hline
\end{tabular}
\end{center}
\caption{Percentage of bad clients blocked by AFA and the average number of training iterations to block them.}
\label{tabDetection}
\end{table*}

\subsection{Robustness and Convergence}
The results of our experimental evaluation are shown in Table~\ref{tabResults}. We considered two cases with 10 and 100 clients for the 4 previously described scenarios. For the \emph{byzantine}, \emph{flipping} and \emph{noisy} scenarios we let there be $30\%$ bad clients, as in \cite{blanchard2017}. In each experiment we run 10 independent training splits, where we split the training data equally across all clients. To assess statistical significance of the results we performed a Wilcoxon Rank-Sum test at the $5\%$ level. Thus, for each experiment in Table~\ref{tabResults} we highlight those results where one or two algorithms perform significantly better than the others (according to the outcome of the statistical test). From the results we can observe that AFA is robust in all scenarios for the 4 datasets, i.e. the test error of the algorithm is not significantly degraded compared to the performance achieved in the \emph{clean} scenarios. As expected, FA's performance is significantly affected under the presence of bad clients, especially in \emph{byzantine} and \emph{flipping} scenarios. It is also interesting to observe that both MKRUM and COMED are very sensitive to label flipping attacks, especially in MNIST, FMNIST and CIFAR-10 for the case of 10 clients. For example, in MNIST, MKRUM and COMED test errors increases up to $78.98\%$ and $11.55\%$ respectively in the \emph{flipping} attack scenario with 10 clients, whereas AFA's error is just $2.96\%$, similar to its performance in the \emph{clean} scenario. MKRUM also offers a worse performance in the \emph{clean} scenario compared to the other algorithms, which is especially noticeable in the case of CIFAR-10, where the error rate is $53.02\%$, i.e. a $20\%$ more than FA's error. This is not the case for COMED, whose performance is similar to AFA and FA when no bad clients are present, except for FMNIST, where COMED exhibits a better performance. However, COMED is not consistently robust to all the scenarios considered in the experiment. 

\begin{figure*}
\begin{center}
\includegraphics[width=\linewidth]{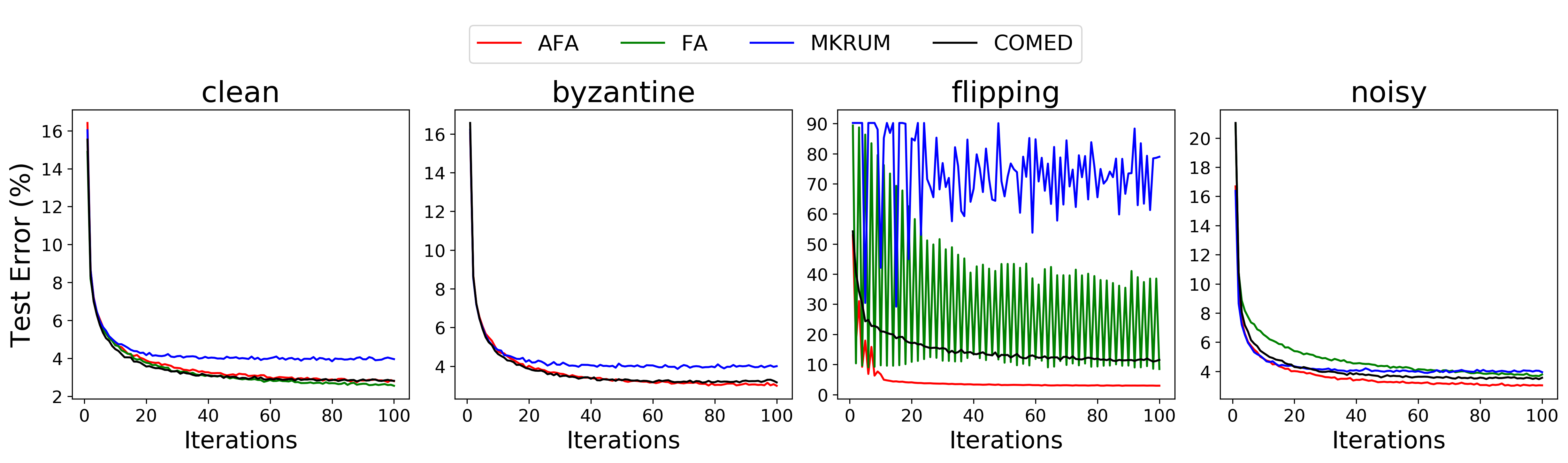}
\end{center}
\caption{Test error (\%) as a function of the number of training iterations for AFA (red), FA (green), MKRUM (blue) and COMED (black) on MNIST with 10 clients for normal operation (all benign clients) and 3 \emph{byzantine}, \emph{flipping}, and \emph{noisy} clients.}
\label{FigConvergence}
\end{figure*}

According to the Wilcoxon Rank-Sum test at the $5\%$ level, AFA outperforms MKRUM with statistical significance in 25 out of the 32 different cases considered\footnote{Note that in Table \ref{tabResults} we just highlight results in bold when one or two algorithms significantly outperform the others.}, whereas MKRUM just outperforms AFA in 1. On the other side, AFA outperforms COMED with statistical significance in 20 cases, whereas COMED just do it in 2. These results support the benefits of AFA, which, in general, offers a better performance than MKRUM and COMED across different scenarios including faulty, noisy and malicious clients.

To analyse the convergence of the algorithms, in Fig.~\ref{FigConvergence} we show the test error as a function of the number of iterations on MNIST with 10 clients. From the results we can observe that AFA converges fast (compared to the other algorithms) to a good solution in all cases. In contrast, MRKUM converges to a worse solution for the \emph{clean}, \emph{byzantine}, and \emph{noisy} scenarios and does not converge for the \emph{flipping} attack. COMED shows a good convergence in all cases except for the \emph{flipping} scenario, where although the algorithm is stable, it achieves a poor performance compared to AFA. The standard implementation of FA performs well on the \emph{clean} case, but it performs poorly for the \emph{byzantine} and \emph{flipping} scenarios. For the sake of clarity, in Fig.~\ref{FigConvergence} we omitted the result for FA in the \emph{byzantine} case as the error is very high (see Table~\ref{tabResults}) compared to the other algorithms. Finally, in the case of having \emph{noisy} clients, we can observe that FA converges slower, as the other algorithms seem to ignore the noisy updates, which helps to speed-up training in the first iterations. The results for all other datasets and scenarios are provided in the supplement.

\subsection{Detection of Bad Clients}
In Table~\ref{tabDetection}, we show the detection rate for bad clients with AFA's Bayesian model with the same experimental settings as described before. In MNIST, FMNIST, and Spambase, AFA is capable of detecting all bad clients in all scenarios. In CIFAR-10, all Byzantine clients are detected and, for label flipping attacks, AFA detects $90\%$ and $100\%$ of malicious clients for the 10 and 100 clients' cases respectively. In contrast, for the noisy scenario, the detection rate is quite low ($20\%$ and $2.7\%$ respectively). However, as shown Table~\ref{tabResults}, for CIFAR-10, AFA's performance is not affected by the presence of noisy clients, when comparing to the performance obtained for the \emph{clean} scenario.

As we set the values of $\alpha_0 = \beta_0 = 3$, for the Beta prior that models the quality of the client's updates, and $\delta = 0.95$ for the threshold to block bad clients, the minimum number of iterations required to block a bad client is 5. As we can observe in Table~\ref{tabDetection}, AFA is capable of detecting the bad clients in a very reduced number of iterations. For example, in most cases, it blocks Byzantine clients in 5 iterations. Label flipping attackers and noisy clients are usually blocked within 10 iterations (except for the noisy clients in CIFAR-10). These results support the accuracy of the aggregation rule in Algorithm~\ref{alg:Aggregation} to detect bad updates and the usefulness of the proposed Bayesian model to block bad clients. For example, in the \emph{byzantine} scenario, from the results in Table~\ref{tabDetection}, AFA would allow to reduce the communications between the clients and the server by $28.5\%$ compared to the other algorithms, as we do not need to send updates to the bad clients once they are blocked.

\subsection{Computational Burden}
\begin{figure}
\begin{center}
\includegraphics[width=7cm]{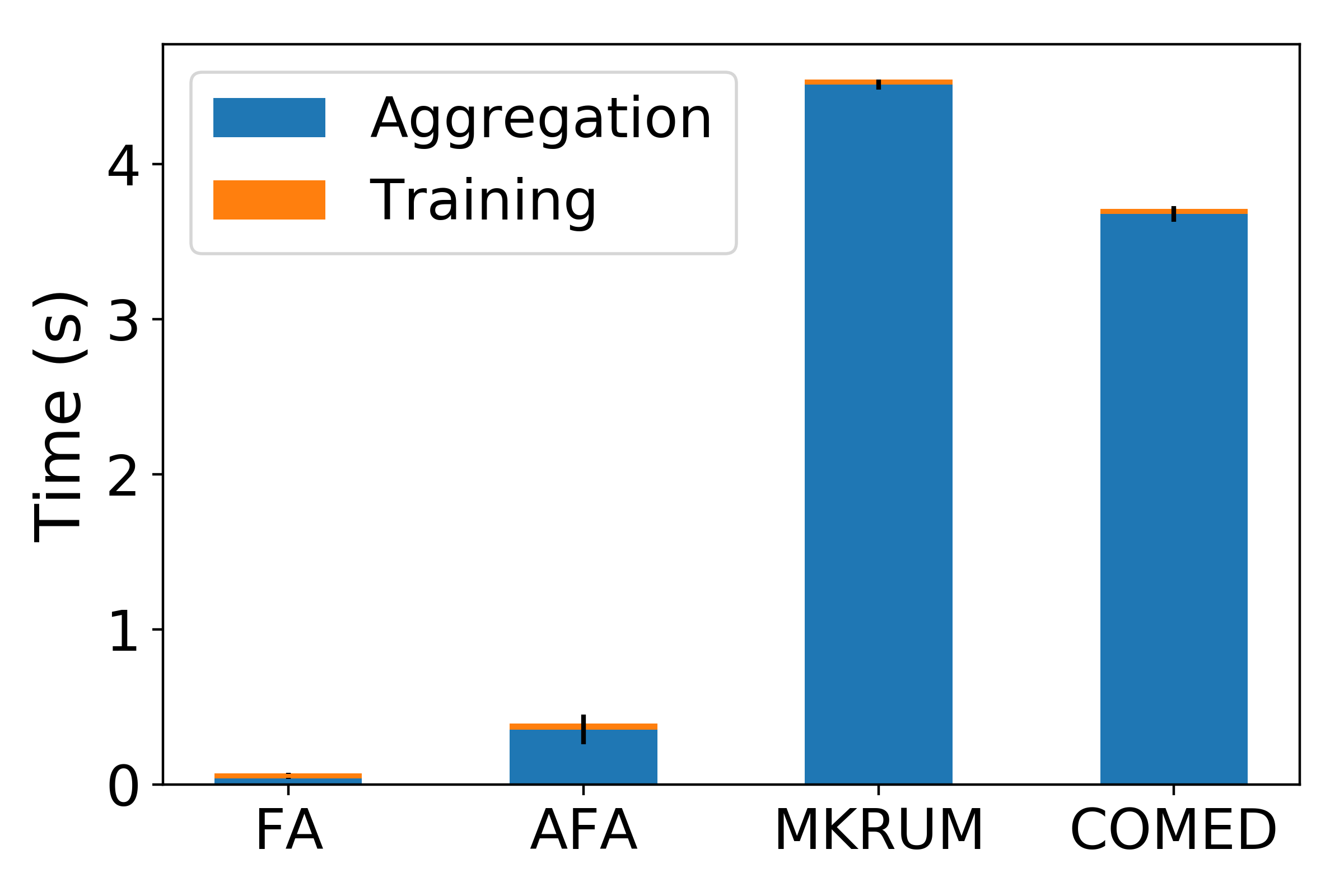} 
\end{center}
\caption{Average training and aggregation time (in seconds) for each iteration on the MNIST dataset with 100 benign clients. Time measured on a standard desktop machine with a GPU nvidia GTX1080 Ti.}
\label{FigTime}
\end{figure}

Finally we performed an experiment to analyse the computational burden of AFA compared to the other methods. For this, we measured both the average time required for the clients to compute the model updates and the average time to aggregate the model in the server. We trained AFA, FA, MKRUM and COMED using MNIST with 100 (benign) clients for 100 training iterations. As in previous experiments, we set the batch size to 200 and, at each iteration, the clients trained the model for 10 epochs. For this experiment, we did not consider the latency in the communications between clients and server. From the results in Fig.~\ref{FigTime} we can observe that AFA is significantly more efficient than both MKRUM and COMED to compute the aggregate model, requiring, on average, $0.35$ seconds to compute the aggregated model as in Algorithm~\ref{alg:Aggregation}, update the Bayesian model and check for possible clients to block. In contrast, MKRUM and COMED required $4.51$ and $3.68$ seconds respectively to compute the aggregate model in the server. As expected, FA, the standard implementation, is the fastest algorithm, requiring just $0.03$ seconds to compute the aggregated model. In the figure, it can be appreciated that, in this case, the difference between the training time and the aggregation time is huge for MKRUM and COMED. In other words, the computational overhead introduced to enhance robustness is very significant when compared to the standard implementation (FA). In contrast, AFA offers a more robust performance (as shown in our previous experiments) with a reduced computational overhead when compared to MKRUM and COMED. It is also important to note that, in the presence of bad clients, the computational burden for AFA would be even lower, as the effective number of clients would be reduced when the malicious ones are blocked. 

\section{Conclusion}
In this paper we presented \emph{Adaptive Federated Averaging} (AFA) an efficient algorithm for Byzantine-robust federated learning. The proposed algorithm is designed to detect and discard bad updates at each iteration, updating the model considering both the fraction of data points provided by each client and the expected quality of the client's model updates. Our algorithm includes a Hidden Markov Model to estimate iteratively the probability of each client providing good model updates. Compared to other robust federated algorithms, this also allows us to identify and block bad clients, improving the robustness and communication efficiency and reducing the computation. 

Our experimental evaluation in 4 real datasets shows that AFA is more robust than other state-of-the-art robust federated learning algorithms, including MKRUM and COMED, when tested against different scenarios with malicious, noisy and faulty clients. For example, we showed that MKRUM and COMED are very sensitive to label flipping attacks, whereas AFA's performance is barely affected. We also showed that AFA is very effective at detecting bad clients and capable of blocking them within a reduced number of training iterations. Finally, our experiments show that AFA is more robust and computationally more efficient than MKRUM and COMED.

Like MKRUM and COMED, our proposed algorithm could still be potentially vulnerable to more subtle or less aggressive attacks. Thus, future research avenues will include the investigation of mechanisms to detect and mitigate more subtle poisoning attack strategies, including targeted attacks \cite{bhagoji2019} and backdoors \cite{bagdasaryan}, which require a finer analysis of the client's updates and their similarities.

\bibliographystyle{aaai}
\bibliography{biblio}

\appendix
\section{Appendix A: Datasets}
In Table~\ref{tabDatasets}, we show the characteristics of the four datasets used in our experiments. For MNIST, FMNIST, and CIFAR-10 we used the standard training and test splits as provided in their respective datasets. For Spambase, we randomly split the dataset with $80\%$ training data points and the rest for testing. 

For MNIST, FMNIST, and CIFAR-10 the inputs are normalized to the interval $[-1, 1]$. In the case of Spambase, we considered the 54 features corresponding to the occurrence of specific keywords in the emails, dropping the last 3 features of the dataset related to the length of sequences of consecutive capital letters. We binarized the selected features, i.e. we only considered the presence or absence of a keyword in an specific email. We used this representation as it led to improved model accuracy and more stability when training the neural network.

\begin{table} [h]
\centering
\begin{tabular}{|l|c|c|c|}
\hline
Name & \# Training & \# Test & \# Features \\
\hline
MNIST & $50,000$ & $10,000$ & $784$ \\
FMNIST & $50,000$ & $10,000$ & $784$ \\
Spambase  & $3,680$ & $921$ & $54$ \\
CIFAR-10 & $50,000$ & $10,000$ & $3,072$ \\
\hline
\end{tabular}
\caption{Characteristics of the 4 datasets used in the experiments.} 
\label{tabDatasets}
\end{table}

\section{Appendix B: Models used in the Experiments}
In Table~\ref{tabArch}, we detail the architectures and the settings used for our experiments. For MNIST, FMNIST, and Spambase we used (fully connected) Deep Neural Networks (DNNs) with 2 hidden layers and Leaky ReLU activation functions. As Spambase is a binary classification problem, we used a Sigmoid activation function for the output layer, whereas for MNIST and FMNIST we used Softmax. For CIFAR-10, we used VGG-11 which is a standard architecture for this dataset. In all cases, we used a gradient descent optimizer and dropout to prevent overfitting. 

\begin{table} [ht]
\centering
\begin{tabular}{|l|}
\hline
{\bf MNIST} and {\bf FMNIST} \\
\hline
- Architecture: DNN ($784 \times 512 \times 256 \times 10$) \\
- Hidden layer act. functions: Leaky ReLU \\ 
\ \ \ \ \ \ \ \ \ \ (negative slope = 0.1) \\
- Output layer act. functions: Softmax \\
- Optimizer: SGD (learning rate = $0.1$, mom. = $0.9$) \\
- Dropout: $p = 0.5$ \\
\hline
\hline
{\bf SPAMBASE} \\
\hline
- Architecture: DNN ($54 \times 100 \times 50 \times 1$) \\
- Hidden layer act. functions: Leaky ReLU \\ 
\ \ \ \ \ \ \ \ \ \ (negative slope = 0.1) \\
- Output layer act. functions: Sigmoid \\
- Optimizer: SGD (learning rate = $0.05$, mom. = $0.9$) \\
- Dropout: $p = 0.5$ \\
\hline
\hline
{\bf CIFAR-10} \\
\hline
- Architecture: CNN (VGG-11) \\
- Optimizer: SGD (learning rate = $10^{-3}$, mom. = 0.9) \\
- Dropout: $p = 0.5$ \\
\hline
\end{tabular}
\caption{Models and parameters used for the 4 datasets in the experiments.}
\label{tabArch}
\end{table}

\section{Appendix C: Convergence}
\begin{figure*}
\begin{center}
\includegraphics[width=6in, trim = 0 0.4in 0 0, clip=true]{figures/by_atk_10u0d.png}
\includegraphics[width=6in, trim = 0 0 0 1in, clip=true]{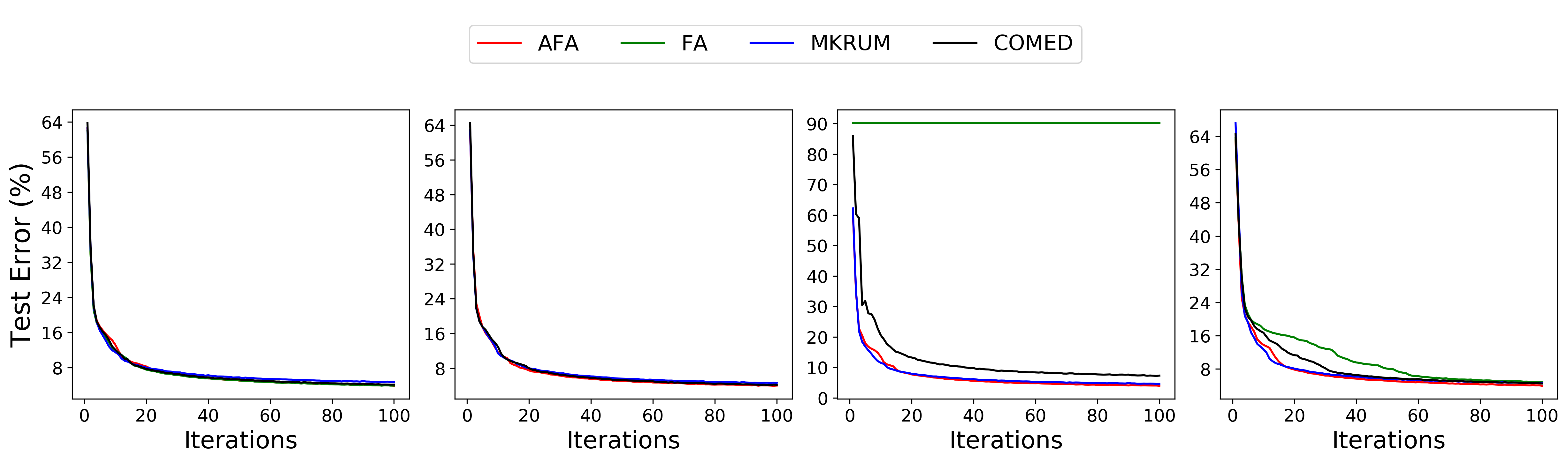}
\end{center}
\caption{Test error (\%) on \textbf{MNIST} dataset as a function of the number of training iterations for the different federated learning methods with (top) 10 clients and (bottom) 100 clients. The \emph{clean} scenario indicates normal operation (all benign clients) while the remaining scenarios, \emph{byzantine}, \emph{flipping}, and \emph{noisy}, all have 30\% faulty, noisy or malicious clients.}
\label{FigConvergenceMNIST}
\end{figure*}

\begin{figure*}
\begin{center}
\includegraphics[width=6in, trim = 0 0.4in 0 0, clip=true]{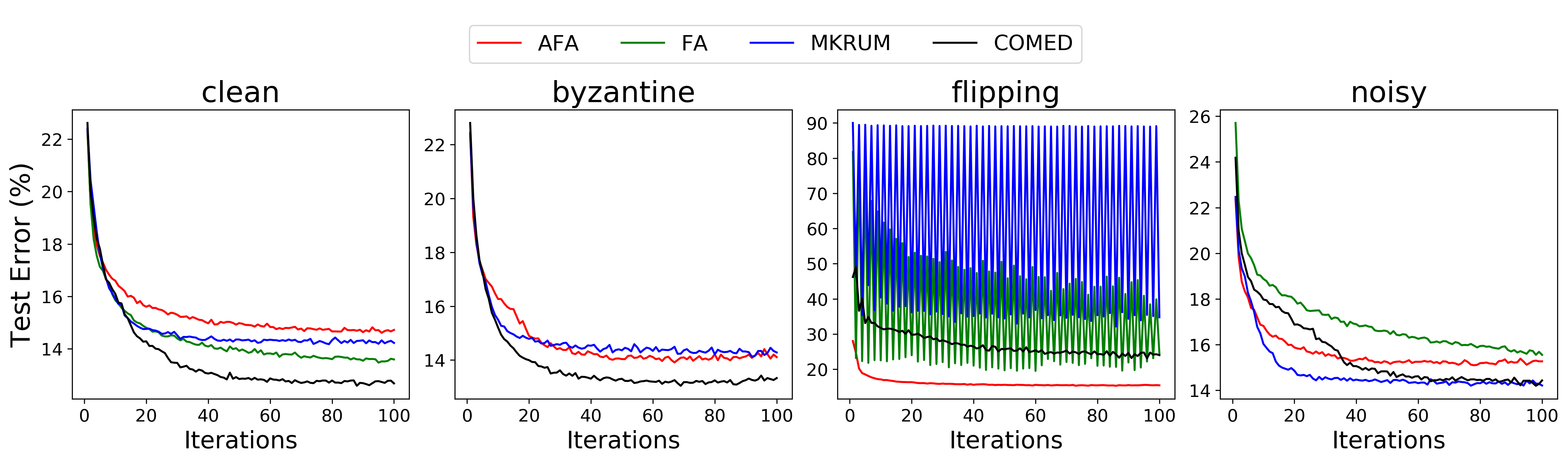}
\includegraphics[width=6in, trim = 0 0 0 1in, clip=true]{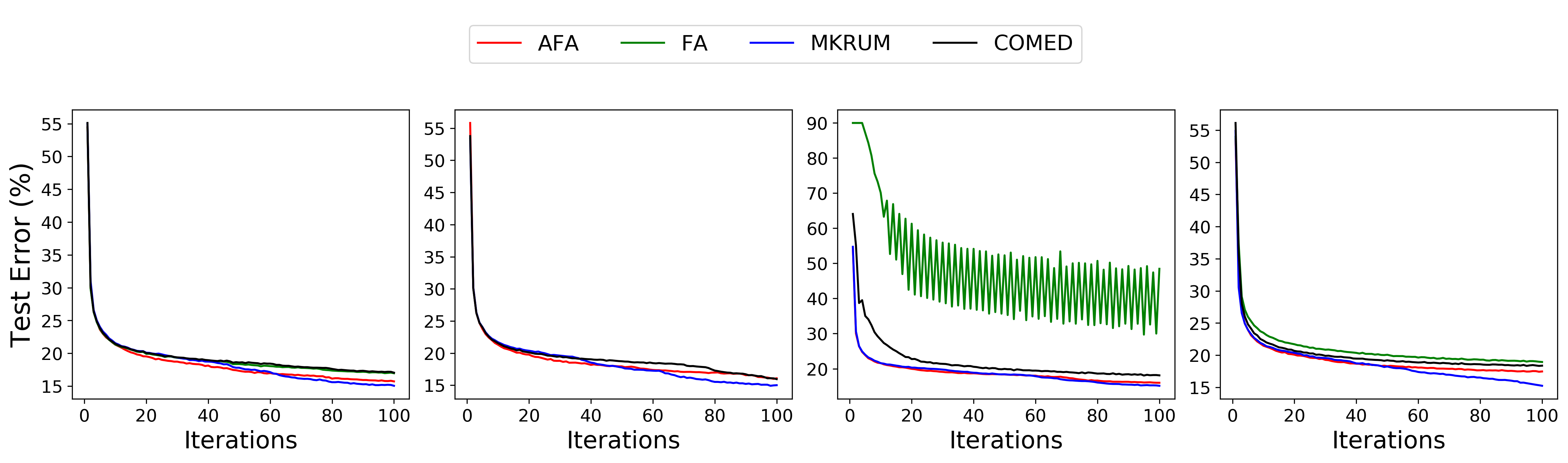}
\end{center}
\caption{Test error (\%) on \textbf{FMNIST} dataset as a function of the number of training iterations for the different federated learning methods with (top) 10 clients and (bottom) 100 clients. The \emph{clean} scenario indicates normal operation (all benign clients) while the remaining scenarios, \emph{byzantine}, \emph{flipping}, and \emph{noisy}, all have 30\% faulty, noisy or malicious clients.}
\label{FigConvergenceFMNIST}
\end{figure*}

\begin{figure*}
\begin{center}
\includegraphics[width=6in, trim = 0 0.4in 0 0, clip=true]{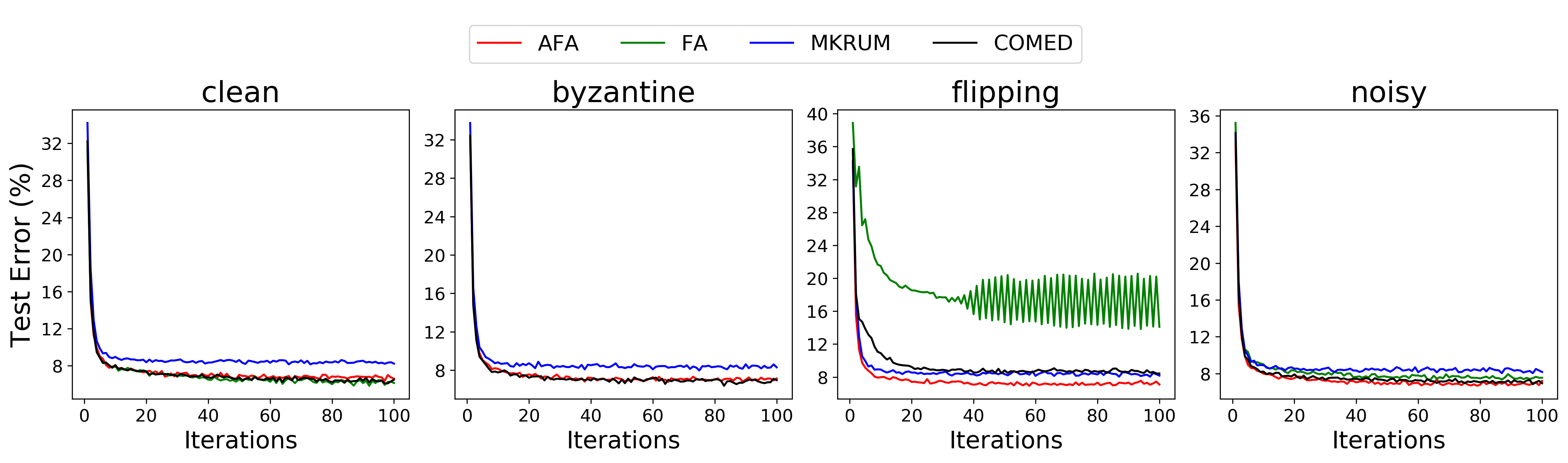}
\includegraphics[width=6in, trim = 0 0 0 1in, clip=true]{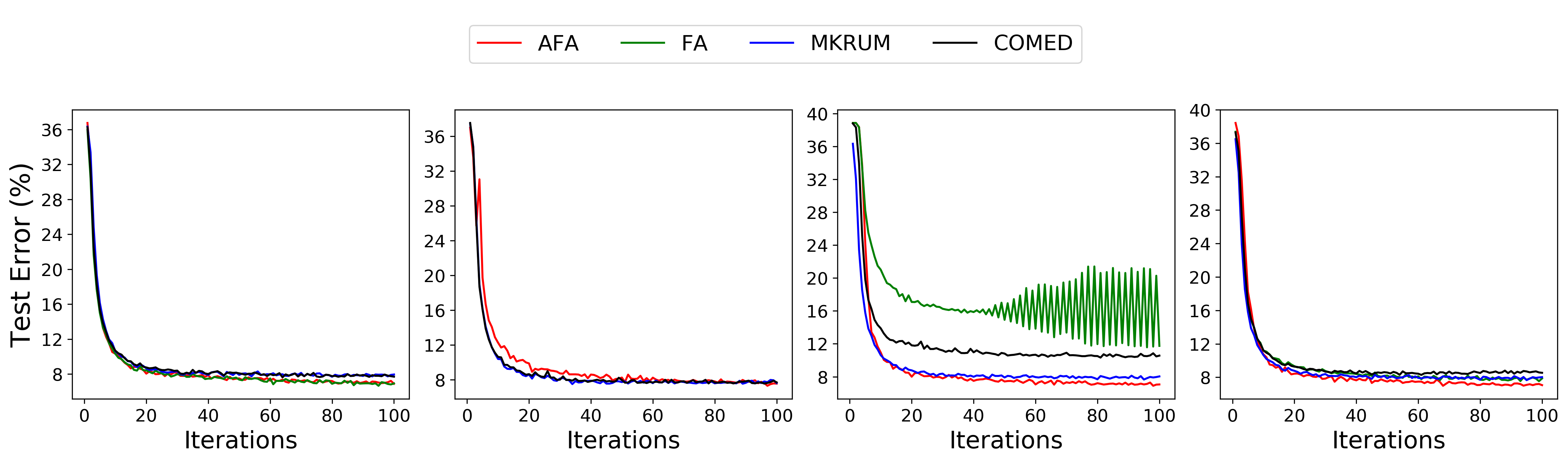}
\end{center}
\caption{Test error (\%) on \textbf{Spambase} dataset as a function of the number of training iterations for the different federated learning methods with (top) 10 clients and (bottom) 100 clients. The \emph{clean} scenario indicates normal operation (all benign clients) while the remaining scenarios, \emph{byzantine}, \emph{flipping}, and \emph{noisy}, all have 30\% faulty, noisy or malicious clients.}
\label{FigConvergenceSpam}
\end{figure*}

\begin{figure*}
\begin{center}
\includegraphics[width=6in, trim = 0 0.4in 0 0, clip=true]{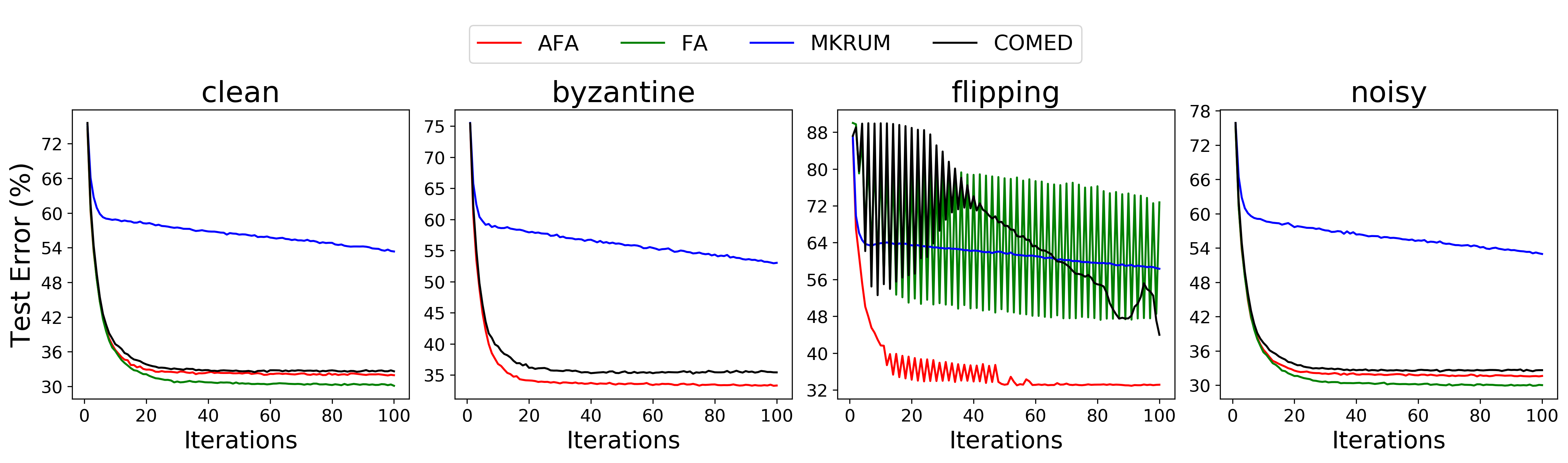}
\includegraphics[width=6in, trim = 0 0 0 1in, clip=true]{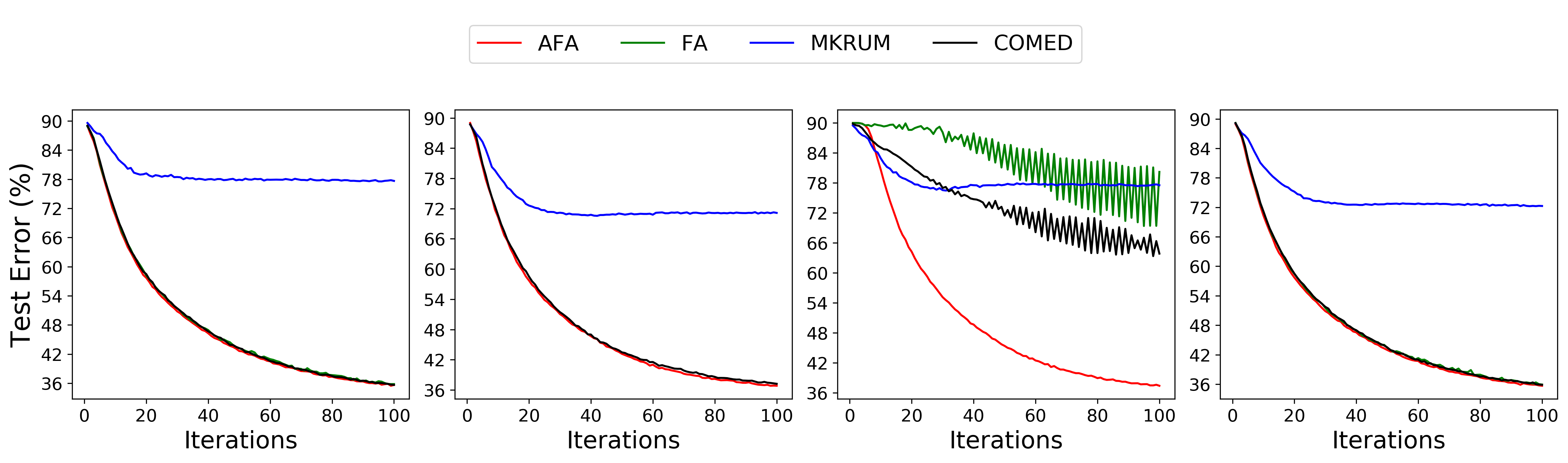}
\end{center}
\caption{Test error (\%) on \textbf{CIFAR-10} dataset as a function of the number of training iterations for the different federated learning methods with (top) 10 clients and (bottom) 100 clients. The \emph{clean} scenario indicates normal operation (all benign clients) while the remaining scenarios, \emph{byzantine}, \emph{flipping}, and \emph{noisy}, all have 30\% faulty, noisy or malicious clients.}
\label{FigConvergenceCIFAR}
\end{figure*}

In Figures~\ref{FigConvergenceMNIST}-\ref{FigConvergenceCIFAR}, we show the classification test error as a function of the number of training iterations for AFA, FA, MKRUM, and COMED over all 4 datasets and all 4 client scenarios considered (\emph{clean}, \emph{byzantine}, \emph{flippling}, and \emph{noisy}) with 10 and 100 clients. We observe that AFA always converges to a stable solution with low test error, whereas the other algorithms are not robust to all scenarios. In the case of MKRUM, we observe notably poor performance on CIFAR-10 when compared to the other methods, as it converges slowly in the case of 10 clients and converges to very high test error in the case of 100 clients even in the \emph{clean} scenario.

The results in these figures are in line with the results presented in the paper, showing the robustness of AFA compared to all other methods. 

\end{document}